\documentclass[runningheads]{llncs}

\usepackage{xspace}
\makeatletter
\DeclareRobustCommand\onedot{\futurelet\@let@token\@onedot}
\def\@onedot{\ifx\@let@token.\else.\null\fi\xspace}

\def\eg{\emph{e.g}\onedot} 
\def\ie{\emph{i.e}\onedot}

\makeatother

\usepackage{graphicx}

\usepackage{amsmath,amssymb} 
\usepackage{color}
\usepackage{epsfig}
\usepackage{graphicx}
\usepackage{subfigure}
\usepackage{booktabs}
\usepackage{multirow}
\usepackage{wrapfig}
\usepackage{placeins}
\usepackage[numbers]{natbib}

\newcommand{\subsubsubsection}[1]{\noindent\textbf{#1}\quad}
\newcommand{\rthree}{\mathbb{R}^3}
\DeclareMathOperator*{\argmin}{arg\,min}

\begin{document}
\pagestyle{headings}
\mainmatter
\def\ECCVSubNumber{Workshop}  

\title{Towards Generalising Neural Implicit Representations} 

\titlerunning{Towards Generalising Neural Implicit Representations}
\authorrunning{T.\ W.\ Costain \& V.\ A.\ Prisacariu}
\author{Theo W.\ Costain\orcidID{0000-0002-7803-6965} \\ Victor A.\ Prisacariu\orcidID{0000-0002-0630-6129}}
\institute{Active Vision Lab, University of Oxford \\ \email{\{costain,victor\}@robots.ox.ac.uk}}

\maketitle
\vspace{-1em}
\begin{abstract}
    \looseness=-1
Neural implicit representations have shown substantial improvements in efficiently storing 3D data, when compared to conventional formats.
However, the focus of existing work has mainly been on storage and subsequent reconstruction.
In this work, we show that training neural representations for reconstruction tasks alongside conventional tasks can produce more general encodings that admit equal quality reconstructions to single task training, whilst improving results on conventional tasks when compared to single task encodings.
We reformulate the semantic segmentation task, creating a more representative task for implicit representation contexts, and through multi-task experiments on reconstruction, classification, and segmentation, show our approach learns feature rich encodings that admit equal performance for each task.
Further, through hold-out experiments, we show that adding semantic supervision when training implicit encoders can significantly improve performance on later unseen tasks, without requiring encoder retraining.
\end{abstract}

\section{Introduction}
\label{sec:intro}
\begin{wrapfigure}{r}{0.49\linewidth}
    \vspace{-1.85em}
    \centering
    \includegraphics[width=0.98\linewidth]{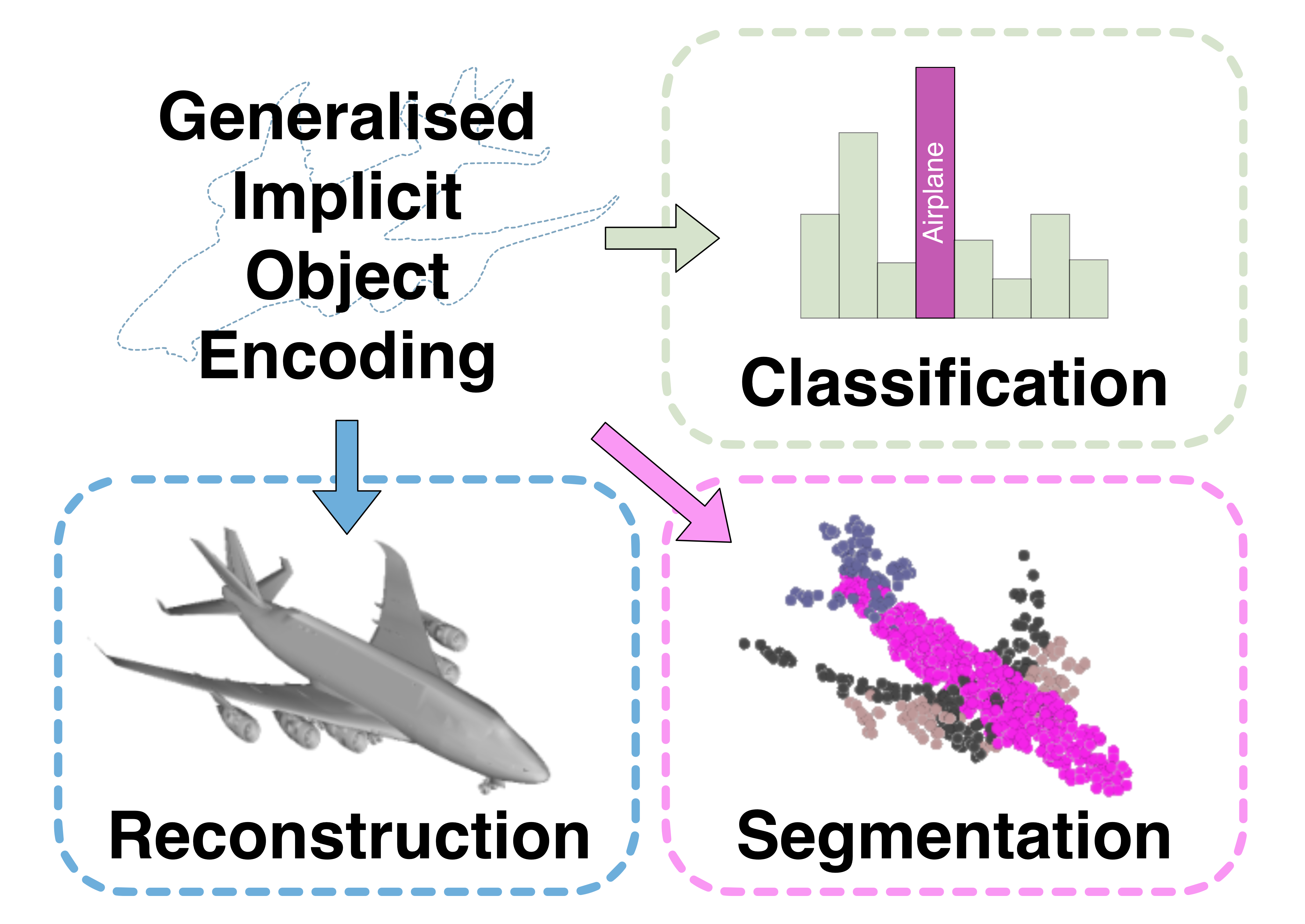}
    \caption{Through multi-task training, implicit representations can be enriched creating a more general representation of a shape or object, and allowing for their use in a number of tasks rather than reconstruction alone.}
    \label{fig:teaser}
    \vspace{-1.5em}
\end{wrapfigure}

Implicit neural representations have garnered significant interest recently for their ability to reconstruct complex 3D structures and 
shapes.
The appeal of these methods stems from a number of useful properties they possess for both reconstructing 3D shapes, as well
as storing them efficiently.
By learning to reconstruct the shapes, networks are able to encode and use a rich set of priors over the 3D domain to improve
the quality of the reconstructions over what can be achieved with classical methods\cite{mescheder2019occupancy, park2019deepsdf}.
Efficiency in storage is achieved by decoupling the encoding from the input and output modality so, unlike voxel based representation,
the storage requirements do not grow cubically with the output resolution.
Further, implicit representations do not suffer from the limitations of mesh and point-cloud based representations, where the quality 
of the reconstruction is typically limited by the output size constraints of a single feed forward pass\cite{mescheder2019occupancy}.

Conditioning a ``decoder'' network on an encoded representation of the input data, neural representations query the network at sample 
point locations for occupancy or distance function information.
This approach allows for reconstructions to be generated with arbitrary resolutions at run-time\cite{mescheder2019occupancy}.

Whilst these properties are impressive, we argue that further useful properties have been left on the table.
In many of the works making use of implicit methods, training is performed with the loss function targeted only at reconstruction accuracy.
This approach, whilst clearly effective, misses a significant potential benefit.
We argue that using a multi task loss, including loss terms related to common tasks such as classification, produces encodings that are
equally effective for reconstruction, but that still provide a richer set of features for use in other downstream tasks.
We suggest that in applications such as augmented reality, where efficient representations are very useful, the ability to encode more 
than just shape information into the representation is likely to be useful.

Whilst a number of works have produced impressive and high quality reconstructions using a number of different approaches, there is a 
general aim in the design neural network encoders to produce features that have meaningful uses beyond a single task.
However, we observe that this aim has not yet translated to implicit representation works.
Many practical applications of implicit representations to real world problems would benefit from the ability to perform multiple
tasks using the same stored data, rather than having to render and re-encode before performing other downstream tasks.

In this paper, we examine the generation of more descriptive neural representation encodings.
Through experiments, we show that encodings generated purely for reconstruction can produce poor results on other tasks.
We demonstrate the expected result that the encodings used in neural representations can be trained to develop properties useful for other
tasks common in computer vision, without any appreciable reduction in reconstruction performance.
We then detail surprising results in our hold-out experiments, showing that training encodings on even a single semantic task significantly improves performance on later held out tasks.
We also argue that the conventional 3D semantic segmentation task does not translate well to implicit representations, where the object being
reconstructed is not or cannot be operated on directly.
To address this, we propose, among other experiments, a re-formulation of the semantic segmentation task that is a more representative
formulation when applied to implicit representations.

In summary, the key contributions of this paper are
\begin{itemize}
    \item Investigation of the use of implicit encodings on a number of common computer vision tasks (in a multi-task setting), showing improved performance in these tasks when compared to reconstruction-only encodings, without compromising reconstruction accuracy.
    \item Showing that the addition of a simple semantic task alongside reconstruction is able to substantially improve performance on new tasks, \emph{without} requiring retraining of the encoder. For example, we show training an encoder for both reconstruction and classification significantly improves later segmentation results.
    \item A re-formulation of the semantic segmentation task, that is more representative of a real world task in the context of implicit representations.
\end{itemize}
\section{Related Work}
\label{sec:red}
Implicit (or Neural) representations have been the subject of much recent work.
Early works focused on single objects\cite{mescheder2019occupancy, sitzmann2019scene, sitzmann2020metasdf, park2019deepsdf, michalkiewicz2019deep, genova2019learning, atzmon2020sal, gropp2020implicit, poursaeed2020coupling, xu2019disn, chen2019learning}, encoding either image or point-cloud input into a feature vector, which is used to condition a decoder network.
These decoder networks typically come in one of two forms: concatenation/biasing or hyper-networks.
In concatenation based conditioning (\cite{dumoulin2018feature-wise} argue that biasing and concatenation are analogous), the encoding is concatenated with the point being queried and then passed through the network.
In hyper-networks, the encodings are passed through a small network, whose outputs are the weights used in the network that predicts the value for a given query point.
The outputs of these decoders can typically be divided into two categories, namely occupancy generating or
signed distance function (SDF) generating\footnotemark.
\footnotetext{Arguably occupancy networks are simply SDF networks with the sign function applied to their output, however this ignores
    the increased complexity in regressing SDF values rather than simply their sign. We discuss this point in more detail in Sec.~\ref{sec:method:occ}}

Early works such as \cite{sitzmann2019scene, park2019deepsdf, mescheder2019occupancy, chen2019learning} showed that simple MLP networks were capable of
representing complex distance functions and occupancy functions.
\cite{park2019deepsdf} (DeepSDF) also detailed the use of auto-decoders to estimate optimal encodings for a given input, using a fixed decoder and simple backpropagation.
\cite{mescheder2019occupancy} (Occupancy Networks) demonstrated the alternative occupancy paradigm for implicit representations, as well as proposing
a procedure to extract high quality meshes in an efficient manner from the implicit representation, using an octree like approach.
Concurrently, \cite{chen2019learning} (IM-Net), also using the occupancy paradigm, showed impressive results for single view reconstruction, as well as both 2D and 3D shape generation.
\cite{michalkiewicz2019deep} learn level sets to represent shapes.
Similarly, \cite{atzmon2020sal} (SAL), used unsigned function priors to train a signed distance function.
\cite{poursaeed2020coupling} combine both explicit atlas based reconstruction and implicit neural reconstruction, enforcing consistency between the two methods.

Later works investigated representing larger scenes\cite{chabra2020deep,Peng2020convoccupancy}.
Many of these methods did not expand the size of the area described by a given embedding, instead proposing methods to recover encodings for a small local region where an implicit representation can then extract local shape information.
\cite{Peng2020convoccupancy} (Convolutional Occupancy Networks) interpolated between encoded points in a volume or plane, to generate the conditioning vector for
a occupancy network in a region around the encoded point.
\cite{chibane2020implicit} took a similar approach, adding also multiple resolutions of encoded volumes.
A separate group of works\cite{jiang2020local, chabra2020deep, tretschk2020patchnets} all took a slightly different approach from above
and divided the scene into regions, generating a small encoding for each region.
Both \cite{jiang2020local} and \cite{chabra2020deep} made use of a grid of small local encodings, whereas \cite{tretschk2020patchnets}
made use of a number of oriented spherical patches of differing radii each with an encoding.

Further improvements to implicit representations in general were proposed by \cite{sitzmann2020implicit} and \cite{tancik2020fourier}
showing that adding higher frequency information to simple networks drastically improved their ability to generate high quality reconstructions.
\cite{duan2020curriculum} proposed a curriculum based learning approach for implicit representations, improving reconstruction quality
of complex local details.

Other works have used implicit representations for a number of other tasks, most notably novel view synthesis\cite{martin2020nerf, mildenhall2020nerf, sitzmann2019deepvoxels, zhi2021inplace}.

There have also been a number of works considering the application of multi-task approaches to 3D problems\cite{pham2019jsis3d, lahoud20193d,hassani2019unsupervised,liang2019multi}.
Multi-task learning has enabled improvements where tasks are related or closely coupled, such as semantic and instance segmentation\cite{lahoud20193d,pham2019jsis3d}.
As well, \cite{hassani2019unsupervised} made use of a multi-task setup in unsupervised training to learn a useful embedding space for 3D point-cloud inputs.

Latent representations have been used to bridge the gap between point-cloud and volumetric representations\cite{meng2019vv}, whilst others have sought to learn directly on meshes rather than point clouds or voxel grids\cite{hanocka2019meshcnn, hu2021subdivision}.

The concurrent work of \cite{zhi2021inplace} examines a similar use of implicit representations in a multi-view synthesis setting (based on \cite{martin2020nerf}).
Their work showed that the addition of geometric information can improve robustness in the semantic task, achieving strong performance in noisy environments while being able to achieve remarkable accuracy with little semantic supervision.
\cite{kohli2020inferring} also examine semantics in a multi-view setting, using \cite{sitzmann2019scene} for the internal representation.
Their method also demonstrates the close relationship between appearance, geometry, and semantics, being able to synthesise novel appearance views from semantic images and vice versa.
\section{Tasks}
\label{sec:method}
In this section, we first cover the principles of implicit representations. We then describe the other tasks we consider, including our variation to
the normal task of segmentation that we use in our experiments, for a fairer representation of the segmentation task in implicit contexts.
\begin{figure*}[t]
    \centering
    \includegraphics[width=0.7\linewidth]{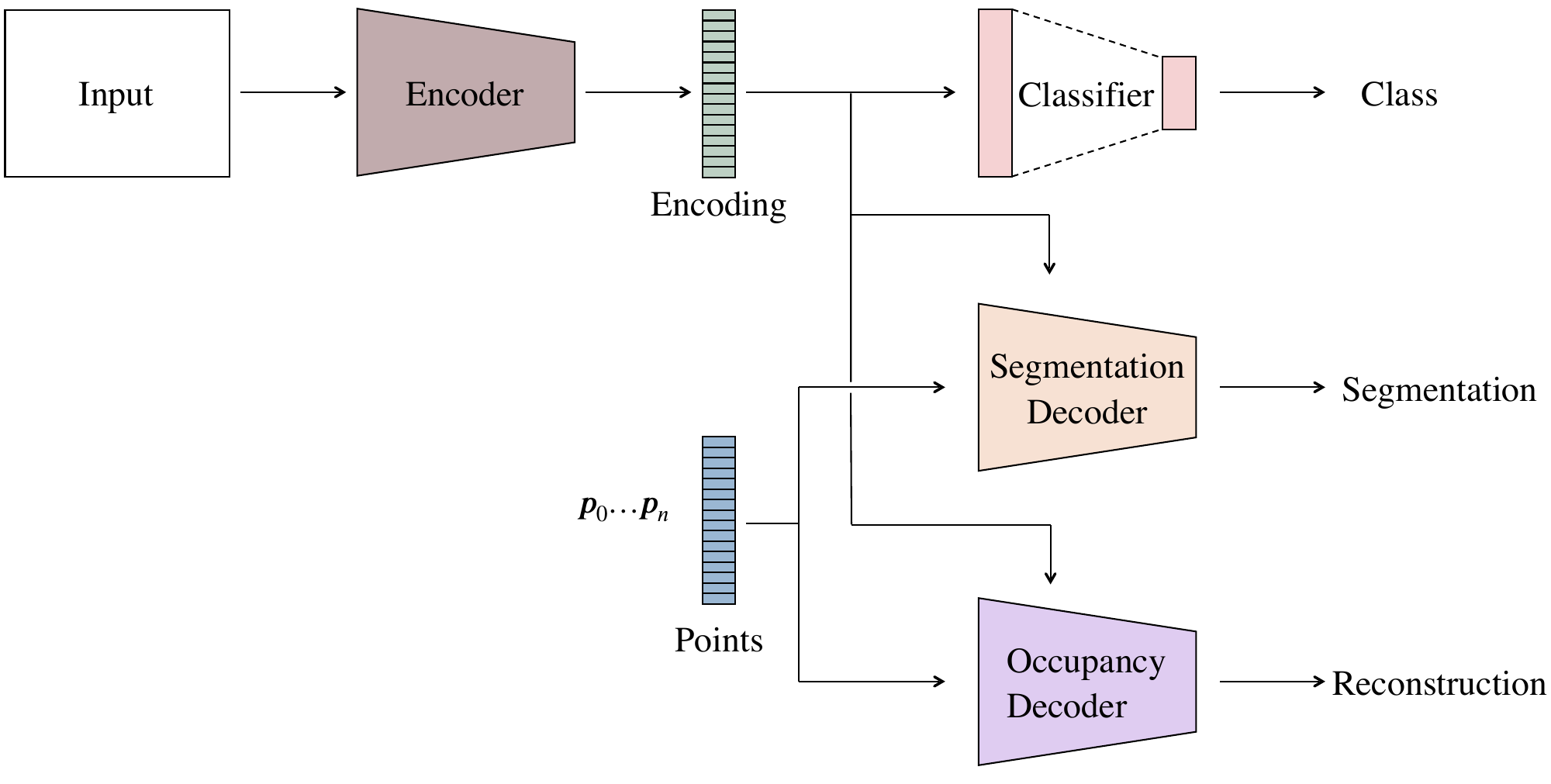}
    \caption{An overview of our network architecture. The network takes as input either images or point-clouds, generating an encoding from them. This
        encoding can then be used in a number of ways. For classification, the encoding is passed directly into a simple classifier. For segmentation and
        reconstruction, the encoding is used to condition the decoder networks. The decoder networks take a number of points as input and returns for each point
        either, the probability that that point lies inside the encoded shape, or semantic label probabilities.}
    \label{fig:arch}
    \vspace{-1.5em}
\end{figure*}

\subsection{Implicit Representation}
\label{sec:method:occ}
Neural implicit representations attempt to estimate the function describing the surface of a given object.
A common formulation is to map from a point in space, $\mathbf{p}\in\rthree$, to the smallest signed distance between the point and
the outer face of a surface, \ie a SDF.
This gives rise to an expression\cite{park2019deepsdf, xu2019disn} of the form $s: \rthree \rightarrow \mathbb{R}$.
Another common formulation is to estimate the probability that a given point lies within the object (\ie probability of occupancy), rather
than regressing the SDF directly.
This gives a function\cite{mescheder2019occupancy} of the form $o: \rthree \rightarrow \{0,1\}$.
However, we note the following relationship
\begin{equation}\label{eq:ots}
    o = \sigma(-s) > \tau
\end{equation}
where $\tau$ is a threshold parameter, and $\sigma$ is a sigmoid function.
This relationship suggests that the occupancy function is a simplified version of the SDF, moving the problem from a regression context
and into a classification context.
Further, this reformulation suggests that where the extra information provided by the SDF (but not the occupancy function) such as the
surface normal (given as, $\frac{\partial s}{\partial\mathbf{p}}$, the spatial gradient of the SDF\cite{park2019deepsdf}) is not needed,
the occupancy function is likely to be easier to learn.
If this is true, we suggest this is the result of the occupancy network only needing to learn a decision boundary over $\rthree$ rather than
having to learn both the boundary and then regress a points distance from it.
In addition, we note that DeepSDF applies clamping to its loss function, such that points away from the surface do not impact the loss.
Further, DeepSDF does not enforce the Eikonal constraint\footnote{A requirement for a ``true'' SDF} like other works\cite{michalkiewicz2019deep, gropp2020implicit}.
Accordingly the network is mainly learning the zero crossing point of the SDF rather than a metric SDF.

This then implies that differences between Occupancy Networks and DeepSDF lie mostly in the training and design of the two networks, and not in the functions they embed.
For this reason, we believe that although the evaluation in this paper is conducted soley on Occupancy Networks, the results should also apply to DeepSDF based encodings.
Given this,  and the simplicity of the method, we chose to use the formulation of Occupancy Networks\cite{mescheder2019occupancy} for our experiments.

\subsection{Other Tasks}
\label{sec:method:seg}
The conventional 3D segmentation task as explored in a number of papers\cite{qi2017pointnet, xie2020review} typically involves predicting a
semantic label for each point in an given input point-cloud.
However, in the context of implicit representations, this task loses much of its meaning, as we want to learn semantic information at locations not in the input pointcloud.
This of particular concern when the input is a degraded and noisy point-cloud (Sec.~\ref{sec:expr}).
As we are considering the occupancy of a given spatial location, it makes more sense to consider the task as determining the semantic
label of regions within the shape.

Hence, given a mesh $\mathcal{X}$, for each vertex $\mathbf{x}\in\mathcal{X}$ with a semantic label $c_\mathbf{x}$, the semantic
class of any location $\mathbf{p} \in \rthree\cap\mathcal{X}$ lying inside the mesh, has the semantic class of the nearest
vertex of $\mathcal{X}$.
\[
    c_{\mathbf{p}} = c_{\mathbf{z}}\qquad\ \text{where}\; \mathbf{z} = \argmin_{\mathbf{x}} ||\mathbf{p} - \mathbf{x}||_2
\]
This scheme has the same effect as producing a Voronoi partitioning of the space inside the mesh.
Points that lie outside the mesh, are considered to be background and therefore have no valid semantic class.
The segmentation task then becomes predicting the label of a point inside the shape according to the nearest neighbour assignment.
During both training and inference, we evaluate the semantic label task at the same locations as the reconstruction task.

As well as segmentation, we also investigate the performance of our approach in the task of classification.
Unlike the segmentation task which requires the implicit code to encode information about the properties of spatial regions (similarly also with the
reconstruction task), classification requires that the encodings allow simple classification networks to discriminate between them.
Later experiments (Sec.~\ref{sec:res:modelnet} \& Sec.~\ref{sec:res:part}), show that the requirements classification has for the encodings are
noticeably different to reconstruction.

Our results show that implicit representations can be encouraged to be more representative of objects, rather than merely encoding their shape.
We focus on two particular tasks that are common tasks, but expect that generalising the encodings over further tasks is likely to also be possible.

\section{Experiments}
\label{sec:expr}
The experiments are divided into three parts.
First we consider the original dataset from \cite{mescheder2019occupancy}, establishing a baseline and some preliminary experiments involving
reconstruction and classification.
Next we examine a more challenging classification task, before turning finally to a semantic segmentation task.

An overview of our network architecture is shown in Fig.~\ref{fig:arch}.
Specific details of the architecture for each experiment are outlined in Sec.~\ref{sec:expr:arch}.
The loss functions used depends on the task.
For the reconstruction loss, $\mathcal{L}_\text{rec}$, we use binary cross entropy as in\cite{mescheder2019occupancy}.
Both classification, $\mathcal{L}_\text{cls}$, and segmentation, $\mathcal{L}_\text{seg}$, use the cross entropy loss.
In the multi task settings, the losses are combined in a weighted linear fashion as
\(
\mathcal{L}_\text{tot} = \mathcal{L}_\text{rec} + \lambda_\text{cls}\times\mathcal{L}_\text{cls}+ \lambda_\text{seg}\times\mathcal{L}_\text{seg}
\).
For all experiments $\lambda_\text{cls}=\lambda_\text{seg}=1.0$.
We use an ADAM optimiser with learning rate of $10^{-4}$.
Training takes approximately 4 days on a NVIDIA GeForce GTX 1080Ti.

\subsection{Datasets}
We perform our experiments on a number of datasets. The original dataset (hereafter Choy) from \cite{mescheder2019occupancy} is the subset of
ShapeNetCore\cite{shapenet2015} from \cite{choy20163d}. We also make use of ModelNet40\cite{wu2015modelnet} for further classification
experiments and ShapeNetPart\cite{yi2016scalable} for our segmentation experiments.
Data pre-processing was accelerated with GNU Parallel\cite{tange2011parallel}.

We limit our experiments to datasets with similar properties to those used in \cite{mescheder2019occupancy}, as we are not seeking to validate
the specific implicit representation format we are using, rather the benefits of more feature rich encodings.
This means that we do not consider larger scale datasets such as Stanford3D\cite{armeni20163d} that our chosen method might struggle with.
We leave this to future experiments with other methods such as \cite{Peng2020convoccupancy} or \cite{chabra2020deep} that are better able to
reconstruct larger scenes.

For all experiments, the properties of the inputs remain constant.
For point-clouds we sample 300 points from the ground truth point-cloud, and apply noise using a Gaussian distribution with zero mean
and standard deviation 0.05 to the sampled point clouds, identically to \cite{mescheder2019occupancy}.
For images we crop and resize the images identically to \cite{mescheder2019occupancy}.

\subsubsection{Choy / ShapeNetCore}
The dataset used in \cite{mescheder2019occupancy} from which our work builds on, uses the renderings and voxelisations\cite{choy20163d}
of a subset of the ShapeNetCore\cite{shapenet2015} dataset. We use the rendered images to train the image based encoder in later
experiments.
The fully processed dataset was provided by \cite{mescheder2019occupancy} as part of their publication.
Briefly, meshes are loaded and a large number of depth images are rendered.
These depth images are fused to form a watertight mesh from which points and their corresponding occupancy value can be sampled.
Although the occupancy samples are not provided as part of the dataset in\cite{choy20163d}, to reduce ambiguity we will refer to the dataset
from \cite{mescheder2019occupancy} as the Choy dataset throughout this paper.
The dataset consists of 30,648 training meshes, 4,358 validation meshes and 3,738 test meshes across 13 object categories.

We use the Choy dataset both for our baseline experiments, as well as some preliminary classification experiments.
Our experiments with this dataset are outlined in Sec.~\ref{sec:res:choy}.

\subsubsection{ModelNet40}
For further classification experiments, we make use of the Model\-Net\-40\cite{wu2015modelnet} dataset.
As rendered images were not readily available, we rendered images using Pyrender\cite{matl2020pyrender} in the same fashion as \cite{choy20163d},
choosing 24 viewpoints with constant radius and altitude, but random azimuth.
The occupancy samples are generated with the code provided by \cite{mescheder2019occupancy}.
The dataset consists of 9,843 training meshes and 2,468 testing meshes across 40 object categories.
Our experiments with this dataset are outlined in Sec.~\ref{sec:res:modelnet}.

\subsubsection{ShapeNetPart}
For our semantic segmentation experiments, we make use of the dataset from \cite{yi2016scalable}, which we refer to as ShapeNetPart.
Again the occupancy samples were generated using the code from \cite{mescheder2019occupancy}.
Semantic labels were assigned to the occupancy samples using the procedure outlined in Sec.~\ref{sec:method:seg} from the ground truth semantic labels
in\cite{yi2016scalable}.
The dataset consists of 12,121 training, 1,854 validation, and 2,858 testing meshes following the corresponding splits from ShapeNetCore.
Our experiments with this dataset are outlined in Sec.~\ref{sec:res:part}.

\subsection{Architecture}
\label{sec:expr:arch}
Our network takes either point-clouds or images as input to the encoder.
In all the experiments, we use the same two encoders: one for point-cloud input, and another for image input.

\subsubsection{Point-cloud input}
We use the same variation on the original network from \cite{qi2017pointnet} as \cite{mescheder2019occupancy}.
In this formulation, the fully connected (FC) layers normally present in the original network are replaced by residual FC
blocks\cite{he2016deep}.
During training the network samples 300 points from the input point cloud and applies Gaussian noise ($\sigma=0.005$) before passing these into the encoder (identically to \cite{mescheder2019occupancy}).

\subsubsection{Image input}
We use a pre-trained ResNet-18\cite{he2016deep}, followed by a linear layer to reduce the output dimension following
\cite{mescheder2019occupancy}.

The encoded features are then passed to a decoder.
For decoding point locations into either occupancy values or semantic labels we use one or more of the following, depending on
the task(s).
For classification, the encoding is passed directly to the classifier.

\subsubsection{Task Decoders}
\subsubsubsection{Occupancy Decoder}
This is the same decoder used in \cite{mescheder2019occupancy}. The network takes a number of points
$\mathbf{p}_0,\mathbf{p}_1,\dots,\mathbf{p}_n$ as input and uses conditional batchnorms\cite{de2017modulating}, which take the encoding as their
input, to condition the network.

\subsubsubsection{Classifier}
A simple 2 layer MLP, that takes the encoding directly as input and returns class probabilities.

\subsubsubsection{Segmentation Decoder}
The same network as the occupancy decoder but with a larger output channel dimension.

\subsubsubsection{Parallel Segmentation and Occupancy Decoders}
This is the configuration shown in Fig.~\ref{fig:arch}, in which the segmentation and occupancy decoders operate in parallel.

\subsubsubsection{Joint Segmentation and Occupancy Decoder}
Also the same network as the occupancy decoder, however rather than two separate networks for each task, the same network performs both tasks
simultaneously.
The output is then sliced along the channel dimension to yield two tensors, one containing the occupancy probability, the other
containing the semantic label probabilities.

\subsection{Encoder Baselines}
\begin{wraptable}{R}{0.49\linewidth}
    \vspace{-3em}
    \centering
    \subtable[Classification on ModelNet40]{
        \resizebox{0.95\linewidth}{!}{
            \begin{tabular}{cc}
                \toprule
                                              & Accuracy      \\
                \midrule
                PointNet\cite{qi2017pointnet} & \textbf{87.1} \\
                ONet Encoder + Classifier     & 85.9          \\
                \bottomrule
            \end{tabular}
        }
    }
    \subtable[Segmentation on ShapeNetPart]{
        \resizebox{0.95\linewidth}{!}{
            \begin{tabular}{cc}
                \toprule
                                                    & mIOU          \\
                \midrule
                PointNet\cite{qi2017pointnet}       & \textbf{83.7} \\
                ONet Encoder + Segmentation Decoder & 83.0          \\

                \bottomrule
            \end{tabular}
        }
    }
    \vspace{-0.5em}
    \caption{Encoder comparison baselines.}
    \label{tab:bl}
    \vspace{-2.2em}
\end{wraptable}
To fairly compare later experiments to Occupancy Networks, we use the encoders from the original paper throughout.
To provide comparisons between our tasks and conventional tasks on point clouds, we run two small experiments to show how the encoder (a modified PointNet) from Occupancy Networks performs compared to a baseline PointNet\cite{qi2017pointnet}.
These experiments are shown in Table~\ref{tab:bl}, using results directly from \cite{qi2017pointnet}.
For classification on ModelNet40, the network receives 1024 input points perturbed by Gaussian noise with zero mean and 0.02 standard deviation, as in \cite{qi2017pointnet}.
For segmentation on ShapeNetPart, the network receives 2048 points as input to the encoder, and the input is also used as the query points ($p_0\dots p_n$ in Fig.~\ref{fig:arch}).
We note that for these experiments the network receives at least 3x more input points, with less additive noise, than in later experiments.

\section{Results}
\label{sec:res}
\subsection{Choy Experiments}
\label{sec:res:choy}
\begin{table}[t]
    \centering
    \begin{tabular}{cccc}
        \toprule
                                       & \multicolumn{2}{c}{Recon.} & Class.                                        \\ \cmidrule(rl){2-3}
                                       & IOU $\uparrow$             & Chamfer L1 $\downarrow$ & Accuracy $\uparrow$ \\
        \midrule
        ONet baseline                  & 0.78                       & 0.0081                  & --                  \\
        Classification baseline        & --                         & --                      & 0.92                \\
        Classification w/ ONet encoder & --                         & --                      & 0.80                \\
        Joint Classification \& ONet   & 0.77                       & 0.0084                  & 0.92                \\
        \bottomrule
    \end{tabular}
    \vspace{0.4em}
    \caption{Experiments on the Choy dataset with point-cloud input, showing shape IOU, Chamfer L1, and classification accuracy.
        For ``Classification w/ ONet encoder'' the encoder is pre-trained on the ONet baseline, fixed, and only the classification decoder trained.
    }
    \label{tab:choy:pc}
    \vspace{-2.0em}
\end{table}

We begin with the dataset from \cite{mescheder2019occupancy}.
Our experiments with point-cloud input are shown in Table~\ref{tab:choy:pc}.
Given the small number of classes and fairly distinct visual properties of the classes in this dataset, the high accuracy in classification is
not unexpected, even with the reduced quality of the input point-clouds.
To evaluate the classification performance of the baseline encoder, we fix its parameters and train a simple 2 layer MLP classifier to operate on its output encodings.
Classification on this fixed encoder shows a substantial reduction in accuracy compared to the jointly trained case, \ie where the encoder is \emph{not} fixed.
Notably in this joint training scenario, full accuracy in both tasks is recovered.

\begin{table}[t]
    \centering
    \begin{tabular}{cccc}
        \toprule
                                       & \multicolumn{2}{c}{Recon.} & Class.                                        \\ \cmidrule(rl){2-3}
                                       & IOU $\uparrow$             & Chamfer L1 $\downarrow$ & Accuracy $\uparrow$ \\
        \midrule
        ONet baseline                  & ( 0.58 / 0.57 )            & ( 0.021 / 0.021 )       & --                  \\
        Classification baseline        & --                         & --                      & ( 0.92 / -- )       \\
        Classification w/ ONet encoder & --                         & --                      & ( -- / 0.63 )       \\
        Joint Classification \& ONet   & ( 0.59 / 0.57 )            & ( 0.020 / 0.023 )       & ( 0.92 / 0.91 )     \\
        \bottomrule
    \end{tabular}
    \vspace{0.4em}
    \caption{Experiments on the Choy dataset with image input, showing shape IOU, Chamfer L1, and classification accuracy. Values in brackets: left with ResNet pre-training, right without.}
    \label{tab:choy:img}
\end{table}
Our experiments with image input are shown in Table~\ref{tab:choy:img}.
The results are similar to the point-cloud experiments. As discussed in \cite{mescheder2019occupancy}, the lower performance in reconstruction for
the ONet can potentially be attributed to occlusion.
We also train the baseline from scratch without ResNet pre-training.
Without pre-training the same ``Classification w/ONet encoder'' experiment shows a similar result to the point-cloud experiment where the network performs poorly on the classification task, as the network is not encouraged to generate features meaningful to other tasks.
Further, without pre-training, the network was significantly less stable, and convergence was slower.
The joint training result shows that the encoding is capable of performing both tasks without loss of accuracy.

\subsection{ModelNet40 Experiments}
\label{sec:res:modelnet}
\begin{table}[t]
    \vspace{-2.0em}
    \centering
    \begin{tabular}{cccc}
        \toprule
                                       & \multicolumn{2}{c}{Recon.} & Class.                                        \\ \cmidrule(rl){2-3}
                                       & IOU $\uparrow$             & Chamfer L1 $\downarrow$ & Accuracy $\uparrow$ \\
        \midrule
        ONet baseline                  & 0.73                       & 0.011                   & --                  \\
        Classification baseline        & --                         & --                      & 0.82                \\
        Classification w/ ONet encoder & --                         & --                      & 0.57                \\
        Joint Classification \& ONet   & 0.70                       & 0.012                   & 0.82                \\
        \bottomrule
    \end{tabular}
    \vspace{0.4em}
    \caption{Experiments on the ModelNet40 dataset with point-cloud input, showing shape IOU, Chamfer L1, and classification accuracy. 
    }
    \label{tab:mn40:pc}
    \vspace{-1.5em}
\end{table}
To better evaluate the classification performance, as well as the shortcomings of the reconstruction encodings in classification, we run the same
experiments as in Sec.~\ref{sec:res:choy} on ModelNet40, a more conventional 3D classification benchmark.

Our experiments with point-cloud input are shown in Table~\ref{tab:mn40:pc}.
The results follow a similar pattern to the point-cloud results from the Choy dataset.
As we expected, when we train the classifier using the fixed encoder from the reconstruction task, the classification performance is poor.
This reduction in performance is much more severe than on the Choy dataset, but is consistent with the increased difficulty shown by the lower
accuracy figure on the classification baseline.
However, this performance loss is completely recovered in the joint training, with only a minor decrease in reconstruction performance.

\begin{table}[t]
    \centering
    \begin{tabular}{cccc}
        \toprule
                                       & \multicolumn{2}{c}{Recon.} & Class.                                        \\ \cmidrule(rl){2-3}
                                       & IOU $\uparrow$             & Chamfer L1 $\downarrow$ & Accuracy $\uparrow$ \\
        \midrule
        ONet baseline                  & ( 0.54 / 0.49 )            & ( 0.034 / 0.038 )       & --                  \\
        Classification baseline        & --                         & --                      & (0.85 / -- )        \\
        Classification w/ ONet encoder & --                         & --                      & ( -- / 0.51)        \\
        Joint Classification \& ONet   & ( 0.51 / 0.48 )            & ( 0.036 / 0.042 )       & ( 0.84 / 0.81 )     \\
        \bottomrule
    \end{tabular}
    \vspace{0.4em}
    \caption{Experiments on the ModelNet40 dataset with image input, showing shape IOU, Chamfer L1, and classification accuracy. Values in brackets: left with ResNet pre-training, right without.}
    \label{tab:mn40:img}
    \vspace{-2.5em}
\end{table}
Our experiments with image input are shown in in Table~\ref{tab:mn40:img}.
Here we see that the joint training is able to recover much of the performance on either of the single tasks.
Again, the experiments without pre-training show similar trends to the fixed encoder point-cloud experiments, where the classifier struggles with the ONet encoder, but full accuracy is recovered in joint training.
We suspect the lower performance of the non-pre-trained networks can be attributed to the more varied nature of ModelNet40.

We also suggest that the reduction in performance for the joint task IOU in Tables~\ref{tab:mn40:pc} \& \ref{tab:mn40:img} is more a result of specifics of the IOU metric than a meaningful reduction in performance.
This can be seen from Table~\ref{tab:choy:img} where the Chamfer L1 loss is reduced by the same amount, but the IOU performance is less affected than in Tables~\ref{tab:mn40:pc} \& \ref{tab:mn40:img}.

\subsection{ShapeNetPart Experiments}
\label{sec:res:part}
\begin{table}[t]
    \centering
    \renewcommand{\arraystretch}{1.3}
    \begin{tabular}{ccccc}
        \toprule
                                                                                      & \multicolumn{2}{c}{Recon.} & Seg.                    & Class.                                \\\cmidrule(rl){2-3}
                                                                                      & IOU $\uparrow$             & Chamfer L1 $\downarrow$ & mIOU $\uparrow$ & Accuracy $\uparrow$ \\
        \midrule
        ONet baseline                                                                 & 0.69                       & 0.010                   & --              & --                  \\
        Classification baseline                                                       & --                         & --                      & --              & 0.95                \\
        Segmentation baseline                                                         & --                         & --                      & 0.53            & --                  \\
        \begin{tabular}{c}Joint Segmentation \vspace{-0.5em}\\\& ONet\end{tabular}    & 0.70                       & 0.0098                  & 0.50            & --                  \\
        \begin{tabular}{c}
            Joint Segmentation \vspace{-0.5em} \\
            \& Classification \& ONet
        \end{tabular}                                         & 0.72                       & 0.0086                  & 0.50            & 0.95                                        \\
        \begin{tabular}{c}Parallel Segmentation \vspace{-0.5em}\\\& ONet\end{tabular} & 0.68                       & 0.011                   & 0.53            & --                  \\
        \begin{tabular}{c}
            Parallel Segmentation \vspace{-0.5em} \\
            \& Classification \& ONet
        \end{tabular}                                      & 0.70                       & 0.0095                  & 0.53            & 0.95                                           \\
        \bottomrule
    \end{tabular}
    \vspace{0.4em}
    \caption{Experiments on the ShapeNetPart dataset with pointcloud input, showing shape IOU, Chamfer L1, segmentation mIOU, and classification accuracy.}
    \label{tab:part:over}
    \vspace{-2.0em}
\end{table}

\begin{table*}[t]
    \centering
    \resizebox{0.99\linewidth}{!}{
        \renewcommand{\arraystretch}{1.8}
        \begin{tabular}{l|ccc|cccccccccccccccc@{\hspace{2em}}c}\toprule
            Decoder Type                       & \rotatebox[origin=c]{90}{ONet} & \rotatebox[origin=c]{90}{Seg.} & \rotatebox[origin=c]{90}{Class.} & \rotatebox[origin=c]{90}{Airplane} & \rotatebox[origin=c]{90}{Bag} & \rotatebox[origin=c]{90}{Cap} & \rotatebox[origin=c]{90}{Car} & \rotatebox[origin=c]{90}{Chair} & \rotatebox[origin=c]{90}{Earphone} & \rotatebox[origin=c]{90}{Guitar} & \rotatebox[origin=c]{90}{Knife} & \rotatebox[origin=c]{90}{Lamp} & \rotatebox[origin=c]{90}{Laptop} & \rotatebox[origin=c]{90}{Motorbike} & \rotatebox[origin=c]{90}{Mug} & \rotatebox[origin=c]{90}{Pistol} & \rotatebox[origin=c]{90}{Rocket} & \rotatebox[origin=c]{90}{Skateboard} & \rotatebox[origin=c]{90}{Table} & Mean        \\[1.5em]\midrule
            \emph{Reconstruction IOU}          & \checkmark                     &                                &                                  & \emph{0.75}                        & \emph{0.71}                   & \emph{0.56}                   & \emph{0.8}                    & \emph{0.70}                     & \emph{0.56}                        & \emph{0.75}                      & \emph{0.7}                      & \emph{0.54}                    & \emph{0.81}                      & \emph{0.53}                         & \emph{0.76}                   & \emph{0.75}                      & \emph{0.73}                      & \emph{0.68}                          & \emph{0.70}                     & \emph{0.69} \\ \midrule
            Seg. Only                          &                                & \checkmark                     &                                  & 0.59                               & 0.46                          & 0.45                          & 0.52                          & 0.63                            & 0.31                               & 0.68                             & 0.55                            & 0.52                           & 0.59                             & 0.53                                & 0.58                          & 0.59                             & 0.39                             & 0.51                                 & 0.57                            & 0.53        \\[1.5em]
            \multirow{2}{*}{Joint Decoder}     & \checkmark                     & \checkmark                     &                                  & 0.55                               & 0.45                          & 0.40                          & 0.51                          & 0.62                            & 0.30                               & 0.66                             & 0.53                            & 0.49                           & 0.59                             & 0.47                                & 0.53                          & 0.56                             & 0.30                             & 0.47                                 & 0.56                            & 0.50        \\
                                               & \checkmark                     & \checkmark                     & \checkmark                       & 0.55                               & 0.42                          & 0.38                          & 0.49                          & 0.62                            & 0.32                               & 0.66                             & 0.54                            & 0.49                           & 0.59                             & 0.45                                & 0.55                          & 0.56                             & 0.3                              & 0.46                                 & 0.56                            & 0.50        \\[1.5em]
            \multirow{2}{*}{Parallel Decoders} & \checkmark                     & \checkmark                     &                                  & 0.59                               & 0.50                          & 0.45                          & 0.53                          & 0.63                            & 0.33                               & 0.68                             & 0.56                            & 0.52                           & 0.59                             & 0.53                                & 0.58                          & 0.60                             & 0.39                             & 0.50                                 & 0.57                            & 0.53        \\
                                               & \checkmark                     & \checkmark                     & \checkmark                       & 0.59                               & 0.51                          & 0.42                          & 0.53                          & 0.63                            & 0.34                               & 0.68                             & 0.55                            & 0.53                           & 0.60                             & 0.55                                & 0.57                          & 0.58                             & 0.38                             & 0.50                                 & 0.57                            & 0.53        \\ \bottomrule
        \end{tabular}
    }
    \vspace{0.4em}
    \caption{Experiments on the ShapeNetPart dataset with point-cloud input, detailing per class results, showing segmentation mIOU. Note the first row is reconstruction IOU.}
    \label{tab:part:detail}
    \vspace{-2.0em}
\end{table*}

Our metric for the segmentation task is mean average Intersection over Union (mIOU).
Points are sampled within the shape and assigned semantic labels by the decoder.
The same sample points are used for both segmentation and reconstruction.
Whilst in a real world scenario points would be sampled both inside and outside the shape, we wish to assess the performance of the segmentation
decoder independently of the reconstruction performance, and so only consider points inside the shape.
The IOU is computed for each part of the shape, and averaged to give a shape IOU.
If there are no ground truth points for a given part (\eg `armrest' is a part of the chair class, but several chair instances do not have armrests),
the part is automatically assigned an IOU of 1.
We can then compute mIOU as the average of the shape IOUs.
At inference time, points are sampled randomly from a padded bounding box of the ground truth object, as in \cite{mescheder2019occupancy}.

We evaluate two different part segmentation decoders: the joint decoder where one decoder produces both segmentation and reconstruction at the same time, and the parallel decoder (the configuration shown in Fig.~\ref{fig:arch}) where two similar decoders handle segmentation and reconstruction respectively.
A more detailed explanation is presented in Sec~\ref{sec:expr:arch}.

\begin{wrapfigure}{r}{0.49\linewidth}
    \centering
    \includegraphics[width=0.95\linewidth]{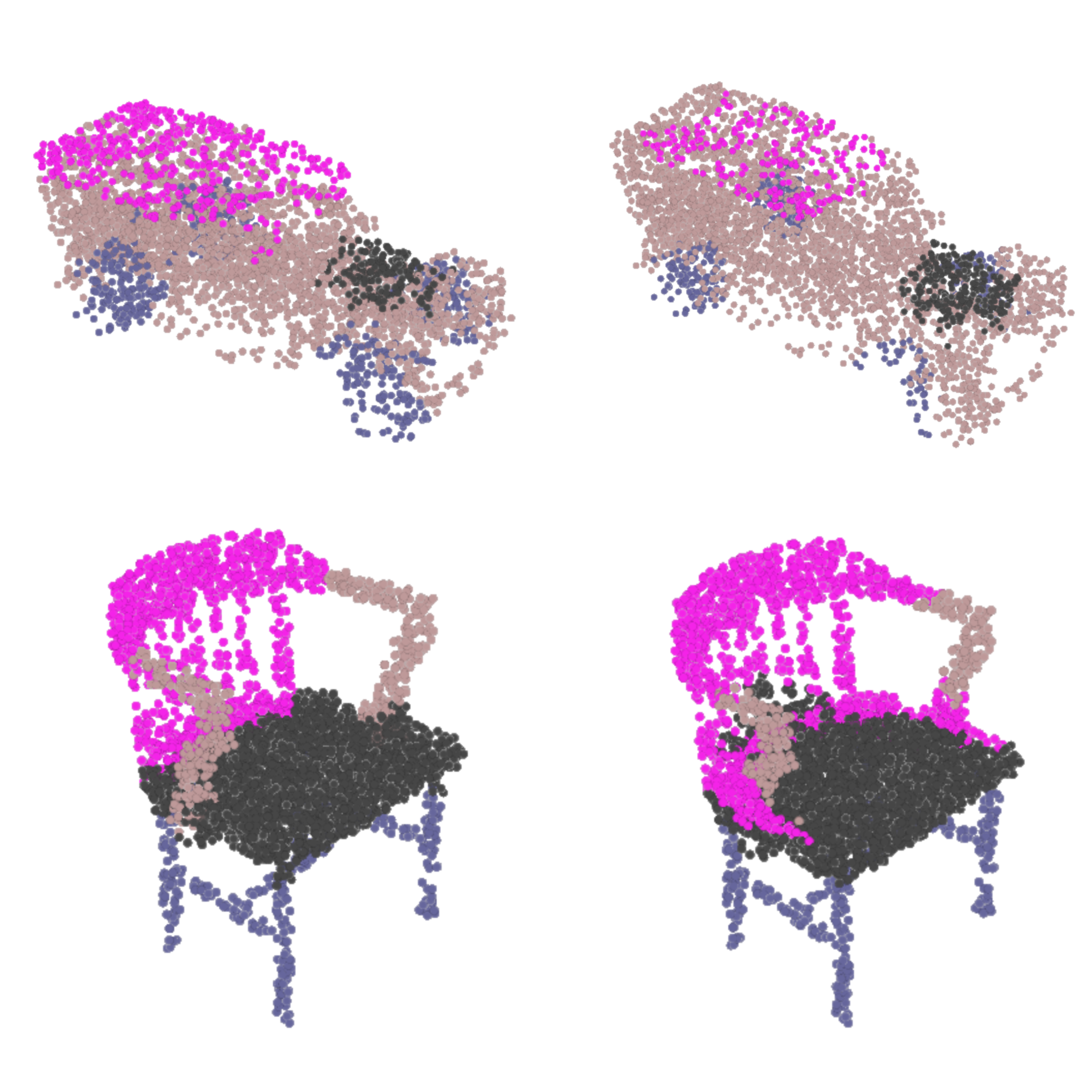}
    \caption{Qualitative joint reconstruction and segmentation results on the ShapeNetPart\cite{yi2016scalable} dataset, showing and example from the car (top) and chair (bottom) classes. Right shows a Joint segmentation \& classification \& ONet reconstructed mesh coloured with semantic class labels. Left shows the same reconstruction with ground truth labels.
    }
    \label{fig:smallqual}
    \vspace{-2.0em}
\end{wrapfigure}
Table~\ref{tab:part:over} shows the reconstruction accuracy, mIOU, and classification accuracy of our different experiments on the ShapeNetPart dataset.
The results show little to no accuracy being lost in any of the tasks for the jointly trained settings.
In the reconstruction task, the network is attempting to learn an encoding that represents the shape properties of a given region of space, such as
the curvature and boundaries.
These properties are likely also useful for the task of segmentation, \ie the semantic class probabilities are potentially somewhat dependant on properties like
local curvature.

Table~\ref{tab:part:detail} shows the per-class segmentation results for the baseline, joint training, and the reconstruction IOU for the baseline.
These results show that the parallel architecture is superior to the joint architecture, and suggest that, although these particular tasks are well correlated at the encoding level, they may be less well correlated within the decoder.
The poor performance on some of the classes such as rocket and headphones may be explained by the thin sections in parts of those objects.
Because the network samples points within the shapes randomly, thin sections like the fins(rocket), cable(earphones), or handlebar(motorbike) are
likely to be undersampled and therefore have poor performance at inference time.
As well as this imbalance, there is also significant imbalance in the number of models in certain categories, which can negatively affect accuracy at inference time.
This is reflected in the higher mIOU scores (across all the experiments), for the classes with more shapes.

Fig.~\ref{fig:smallqual} shows selected qualitative joint reconstruction \& segmentation results.

\subsection{Hold-out Experiments}
Table~\ref{tab:ablation} outlines our hold-out experiments, in which we aim to show the task generalisation of the network.
In these experiments, we train the encoder and reconstruction decoder alongside either the segmentation decoder or the classifier. These weights are then frozen and the remaining decoder is trained.

As the above results show, an encoder trained for reconstruction alone learns features that encode shape information well.
However the results suggest that shape information alone is not sufficient information for either segmentation or classification where significant accuracy is recovered for both tasks in the joint setting.
Table~\ref{tab:ablation} shows a particularly interesting result in that the joint training obtains similar results regardless of whether classification or segmentation was used alongside reconstruction to train the encoder.
We suggest the following explanation for this behaviour: the reconstruction task teaches  geometric information to the network, but not necessarily any semantic information.
\begin{table}[t]
    \centering
    \renewcommand{\arraystretch}{1.3}
    \begin{tabular}{cccc}
        \toprule
                                                          & Recon.         & Seg.            & Class.              \\
                                                          & IOU $\uparrow$ & mIOU $\uparrow$ & Accuracy $\uparrow$ \\
        \midrule
        Classification w/ ONet encoder                    & --             & --              & 0.89                \\
        \begin{tabular}{c}Segmentation \vspace{-0.5em} \\
            w/ ONet encoder\end{tabular} & --             & 0.48            & --                                       \\
        \begin{tabular}{c}Multi Task\vspace{-0.5em} \\
            w/ Seg \& ONet encoder\end{tabular}    & 0.70           & 0.53            & 0.95                           \\
        \begin{tabular}{c}Multi Task \vspace{-0.5em} \\
            w/ Classification \& ONet encoder\end{tabular}   & 0.70           & 0.53            & 0.96                 \\
        \bottomrule
    \end{tabular}
    \vspace{0.5em}
    \caption{Hold-out experiments on the ShapeNetPart dataset with point-cloud input, showing shape IOU, segmentation mIOU, and classification accuracy.}
    \label{tab:ablation}
    \vspace{-1.8em}
\end{table}
Both segmentation and classification are able to provide sufficient semantic information that, when combined with the geometric information, give sufficient capability for new tasks that require semantic and/or geometric information.
To support this we point out that training for classification and reconstruction, despite providing no further point specific information than reconstruction only, still improves the point specific segmentation task by 5\%.
Furthermore, training the encoder for segmentation and reconstruction, despite providing no explicit class information, allows the network to recover the full accuracy on the later classification task, giving a performance improvement of 6\%.
We note that, given this dataset is a fairly simple and quite biased as a classification task, an improvement of this amount suggests noticeably degraded performance (for the reconstruction-only encoder) on this task.


\section{Conclusion}
\label{sec:conc}
In this paper we have discussed generalising the encodings used by implicit representations to a broader range of tasks.
We discuss the current narrow focus of implicit representations, and the potential issues this raises for applications of implicit representations in the real world.
We introduced a modified formulation of the conventional segmentation task that is more applicable to implicit contexts, and detail an appropriate
network to use for this new formulation.
We show that as currently used, implicit encodings struggle to match the performance of the original data they aim to replace on common computer vision tasks, such as classification or part segmentation.
We choose two common tasks and demonstrate that through multi-task training, we can enrich the encodings, achieving strong performance
across the tasks without any loss in reconstruction accuracy.
Through hold-out experiments, we showed improved performance on unseen tasks, when compared to single task training, without needing to retrain the encoder.

\section*{Acknowledgements}
This work was supported by the Engineering and Physical Sciences Research Council [EP/R513295/1]

\FloatBarrier

\FloatBarrier

{\small
    \bibliographystyle{splncs04}
    \bibliography{bibliography}
}

\end{document}